\documentclass[conference]{IEEEtran}
\IEEEoverridecommandlockouts
\usepackage{cite}
\usepackage{amsmath,amssymb,amsfonts}
\usepackage{algorithmic}
\usepackage{graphicx}
\usepackage{textcomp}
\usepackage{xcolor}
\usepackage{multirow}
\usepackage{hyperref}

\def\BibTeX{{\rm B\kern-.05em{\sc i\kern-.025em b}\kern-.08em
    T\kern-.1667em\lower.7ex\hbox{E}\kern-.125emX}}

\DeclareRobustCommand*{\IEEEauthorrefmark}[1]{%
  \raisebox{0pt}[0pt][0pt]{\textsuperscript{\footnotesize #1}}%
}

\usepackage{fancyhdr}
\begin{document}

\title{Analyzing the contribution of different passively collected data to predict Stress and Depression}

\author{\IEEEauthorblockN{Irene Bonafonte\IEEEauthorrefmark{1,7}, Cristina Bustos\IEEEauthorrefmark{1,4}, Abraham Larrazolo\IEEEauthorrefmark{2,4}, Gilberto Lorenzo Martínez Luna\IEEEauthorrefmark{2},  \\ Adolfo Guzm\'an Arenas\IEEEauthorrefmark{2}, Xavier Bar\'o\IEEEauthorrefmark{1}, Isaac Tourgeman\IEEEauthorrefmark{3}, Mercedes Balcells\IEEEauthorrefmark{4,6} and Agata Lapedriza\IEEEauthorrefmark{1,5}}

\IEEEauthorblockA{\IEEEauthorrefmark{1} Universitat Oberta de Catalunya, Barcelona, Spain
}
\IEEEauthorblockA{\IEEEauthorrefmark{2} Instituto Politecnico Nacional, CDMX, Mexico}
\IEEEauthorblockA{\IEEEauthorrefmark{3} Albizu University, Miami, FL, USA
}
\IEEEauthorblockA{\IEEEauthorrefmark{4} Massachusetts Institute of Technology, Cambridge, MA, USA \\
}
\IEEEauthorblockA{\IEEEauthorrefmark{5} Northeastern University, Boston, MA, USA
}
\IEEEauthorblockA{\IEEEauthorrefmark{6} IQS School of Engineering, Universitat Ramon Llull, Barcelona, Spain \\
}
\IEEEauthorblockA{\IEEEauthorrefmark{7} Helmholtz Munich, Neuherberg, Germany \\
}

Email: irene.bonafonte@helmholtz-munich.de, mbustosro@uoc.edu, alarrazolob2021@cic.ipn.mx, lluna@cic.ipn.mx, \\
aguzman@ieee.org,  xbaro@uoc.edu, itourgeman@albizu.edu, merche@mit.edu, alapedriza@uoc.edu
}

\maketitle
\thispagestyle{fancy}

\begin{abstract}

The possibility of recognizing diverse aspects of human behavior and environmental context from passively captured data motivates its use for mental health assessment. In this paper, we analyze the contribution of different passively collected sensor data types (WiFi, GPS, Social interaction, Phone Log, Physical Activity, Audio, and Academic features) to predict daily self-report stress and PHQ-9 depression score. First, we compute $125$ mid-level features from the original raw data. These $125$ features include groups of features from the different sensor data types. Then, we evaluate the contribution of each feature type by comparing the performance of Neural Network models trained with all features against Neural Network models trained with specific feature groups. Our results show that WiFi features (which encode mobility patterns) and Phone Log features (which encode information correlated with sleep patterns), provide significative information for stress and depression prediction. 
\end{abstract}


\section{Introduction}

Wearable devices and smartphones have enabled the collection of large amounts of data that reflect human behavior patterns. This has given rise to a field called Digital Phenotyping, which attempts to quantify behavior using passively captured mobile data \cite{perez2021wearables}. This field has gained interest in mental health diagnosis due to its ability to capture behaviors and habits like sociability, physical activity, mobility, sleep, among others \cite{moura2022digital}. 
Mental health conditions are highly prevalent. However, the diagnosis and management of these conditions face limitations like barriers in seeking help or reliance on self-reports during short clinical visits \cite{williams2017undiagnosed}. Hence, Digital Phenotyping can be a potential tool to support decisions in a clinical setting.

Extensive research in Digital Phenotyping aims to predict mental states or mental disorders, but there are still a lot of open questions. One of them is understanding well what type of data is actually relevant for mental health assessment. From the Machine Learning perspective, most of the solutions do not provide sufficient information about which factors or what types of features contribute to the model's inference of certain mental state or disorder. As a consequence, it is difficult for mental health specialists to rely on model's predictions to evaluate or intervene in patient diagnosis \cite{moura2022digital}. 

In this work, we study the contribution of different data types, coming from different sensors and sources of information, like GPS, WiFi, or phone logs for predicting mental health states using deep learning models. Concretely, we focus on two particular tasks: prediction of the depression score and prediction of the self-report stress level. 
We use the StudentLife dataset \cite{wang2014studentlife} for this study. StudentLife is one of the few public and available datasets for Digital Phenotyping. 

Inspired by the feature analysis done in \cite{umematsu2019improving}, we study the importance of features by analyzing the performance of a neural network trained with all the features extracted from passive sensing data or trained just using a group of features. We can gain insight into which features are more relevant for predicting depression or stress. 
Until now, we have not found a previous study that evaluates the importance of features for prediction of the depression score and prediction of the stress level using the StudentLife dataset. In our analysis, we find the WiFi features, which represent the mobility patterns of a student, are the most discriminant for the neural models for both prediction tasks. Additionally, phone log features show higher discrimination information for stress level classification.
For reproducibility purposes, our feature extraction and models can be found in our public code repository\footnote{\url{https://github.com/cristinabustos16/digitalPhenotyping_featuresAnalysis}}. 




\subsection{Related Work}

Our study is conducted on the StudentLife dataset \cite{wang2014studentlife}, which includes data from 48 students from Dartmouth College passively monitored during a 10-week academic term. The dataset contains passive sensing data from different sensors and an Ecological Momentary Assessment (EMA) component that provides self-reported data such as clinical survey responses, daily stress, sleep hours, mood, and exercise, among others.
The StudentLife dataset has been widely used to study depression diagnosis 
\cite{kim2021automatic,colbaugh2020detecting}
sedentary behavior prediction and activity recommendation \cite{rojas2020activity}, 
food purchase prediction \cite{chen2014my}, recommendations for improving academic performance \cite{djeghri2021recommendation}, 
mood monitoring \cite{ma2021health}
and stress level prediction 
\cite{shaw2019personalized,acikmese2019prediction,kadri2022hybrid}

Stress level prediction on the StudentLife dataset has been approached in different studies. It was first explored in \cite{mikelsons2017towards}, authors categorized the stress level into below-median, median or above-median based on the user’s median, and use a Fully Connected Network with mobility and temporal features.
Later, Shaw et al. \cite{shaw2019personalized} utilized 24-hour histograms of passive data and Long-Short-Term-Memory (LSTM) autoencoder to transform temporal data into a vector representation, then trained a personalized multi task neural network
on a reduced sample of 23 students. Furthermore, stress prediction has been addressed as a binary classification problem, as seen in \cite{acikmese2019prediction}, where authors used LSTMs on the last 2-12h of sensor data to predict between stress / not stress.

Predicting depression using the StudentLife dataset has been addressed by dividing the depression score into two classes (depressed/not depressed). For example, in \cite{gerych2019classifying,saeb2016relationship} authors computed features from the GPS sensor data (e.g. location variance, speed mean, total distance, cluster of locations, circadian movement, entropy and heuristic that indicates the time spent at home). Saeb et al. \cite{saeb2016relationship} found the mobility features grouped in blocks of 2-weeks are significantly correlated with the depression score. In \cite{gerych2019classifying}, authors 
extract mobility features with an autoencoder and then use an SVM classifier on top of these features. Colbaugh et al. \cite{colbaugh2020detecting} used GPS, WiFi, screen lock, light sensor, and microphone data, 
while Kim et al. \cite{kim2021automatic} used the screen lock and unlock data for the last 2-weeks.

\section{From Passive Data to Features}
\label{features}


Inspired by the hierarchical framework introduced by Mohr et al. \cite{mohr2017personal}, which transforms raw data into informative mid-level features (e.g. location, exercise, bedtime, phone usage patterns, and social activity), we compute a collection of $125$ informative features, representing the student's daily information divided into three periods as defined by Wang et al. \cite{wang2014studentlife}: day (9 am - 6 pm), night (12 am - 9 am), and evening (6 pm - 12 am). We divided the features into groups according to the sensor or/and behavior patterns representation. We excluded data contained in the EMA like self-reported sleep hours,  exercise patterns, because we analyze the model performance mostly using passive data. The rest of this section describes the features we computed:

\textbf{WiFi} (36 features): WiFi data tracks the student's connection time to different locations on the campus (over 60 different locations), reflecting their mobility patterns.  
We compute two summary variables: the number of different locations visited and the variance in time spent at each location. Additionally, we include 3 variables representing the time spent in the student's three most visited locations, typically reflecting their homestay and classroom building. Furthermore, we include 7 variables that indicate the time spent in the 7 most frequently visited locations for all students.

\textbf{GPS} (30 features): From the GPS data, we compute and aggregate the maximum distance, the total distance, distance variance, mean speed, speed variance, area of the convex hull that contains all the visited coordinates, and the total time spent indoors and outdoors. Notice that this information is complementary to just identifying frequented locations, which are already encoded with the WiFi features. 

\textbf{Social} (9 features): We computed number of SMS, calls, app usages, Bluetooth contacts, and the total time in seconds the student spend in a phone conversation. 

\textbf{Phone Log} (14 features): These features provide information about the duration in seconds of the smartphone in certain state: phone charging time, phone lock time, and light sensor (time in a dark environment).

\textbf{Activity} (12 features): These features encode the time spent doing physical activity: running, walking, and stationary. 

\textbf{Audio} (9 features): These features represent the time the student spent in silent, conversational or noisy environment. 

\textbf{Academic} (13 features): We include the academic performance information (GPA), the interaction with the campus website like the number of views, contributions, questions, notes and answers. We also add variables such as days to the nearest deadline, daily class hours, and the day of the week.

\section{Neural Network Architectures}

\begin{table*}[htbp]
\caption{Stress Level Classification. Results (F1 and accuracy) for  Binary L-H , Binary LM-H, and Multiclass with FCN -top- and LSTM -bottom- models. Best and second best results per each model (row) and marked in bold face.}
\begin{center}
\begin{tabular}{|c|c||c|c||c|c|c|c|c|c|c|}
\hline
\multicolumn{11}{|c|}{\textbf{FCN}} \\
\hline
  & metric & most freq. & all features & WiFi & GPS & Social & phone log & Activity & Audio & Academic \\
\hline
\multirow{ 2}{*}{L-H} & F1 & 48,3  & \textbf{61,1}$\pm$ 3,5 & 53,7 $\pm$ 3,9 & 46,8 $\pm$ 3,6 & 45,6 $\pm$ 3,2 & \textbf{55,6} $\pm$ 3,9 & 50,6 $\pm$ 2,9 & 50,4 $\pm$ 3,4  &  49,0 $\pm$ 2,7\\
& Acc. &  52,6 & \textbf{60,8} $\pm$ 3,2 & 53,4 $\pm$ 3,9 & 46,8 $\pm$ 3,4 & 45,5 $\pm$ 3,5 & \textbf{55,4} $\pm$ 3,8 & 50,4 $\pm$ 3,2 & 50,1 $\pm$ 3,6  &  48,8 $\pm$ 2,6\\
\hline
\multirow{ 2}{*}{LM-H}& F1 & 38,8  & \textbf{61,4} $\pm$ 3,8 & 55,8 $\pm$ 2,9 & 52,5 $\pm$ 3,2  & 50,6 $\pm$ 3,7 & \textbf{56,8} $\pm$ 3,0 & 53,7 $\pm$ 3,6 & 55,8 $\pm$ 3,6 & 52,4 $\pm$ 2,9 \\
& Acc. & 40,1  & \textbf{64,7} $\pm$ 3,0 & 59,8 $\pm$ 2,4 & 58,6 $\pm$ 1,8 & 56,1 $\pm$ 2,6 & \textbf{60,5} $\pm$ 2,4 & 58,8 $\pm$ 2,7 & 59,9 $\pm$ 2,4  &  58,6 $\pm$ 2,1\\
\hline
\multirow{ 2}{*}{Multiclass}  & F1 & 24,2 & \textbf{50,9} $\pm$ 3,2 & \textbf{43,8} $\pm$ 3,9 & 43,7 $\pm$ 2,4 & 40,0 $\pm$ 2,8 & 43,5 $\pm$ 2,5 & 43,5 $\pm$ 2,6 & 43,6 $\pm$ 2,3  & 42,9 $\pm$ 3,5  \\
& Acc. & 32,3  & \textbf{51,9} $\pm$ 3,2 & \textbf{44,9} $\pm$ 2,4 & 44,8 $\pm$ 2,4 & 41,3 $\pm$ 2,9 & 44,8 $\pm$ 2,5 & 44,5 $\pm$ 3,0 & 44,5 $\pm$ 2,3  &  44,1 $\pm$ 3,4\\
\hline
\hline
\multicolumn{11}{|c|}{\textbf{LSTM}} \\
\hline
  & metric & most freq. & all features & WiFi & GPS & Social & Phone log & Activity & Audio & Academic \\
\hline
\multirow{ 2}{*}{L-H} & F1 & 48,3  & \textbf{61,4} $\pm$ 5,0 & \textbf{57,5} $\pm$ 2,9 & 49,7 $\pm$ 2,9 & 48,8 $\pm$ 2,9 & 54,5 $\pm$ 4,9 & 51,8 $\pm$ 2,7 & 49,0 $\pm$ 3,1  &  48,8 $\pm$ 2,8\\
& Acc. & 52,6  & \textbf{61,5} $\pm$ 2,8 & \textbf{57,8} $\pm$ 2,7 & 49,7 $\pm$ 3,4 & 49,1 $\pm$ 3,0 & 55,9 $\pm$ 4,3 & 52,7 $\pm$ 3,0  & 48,7 $\pm$ 3,2  &  48,8 $\pm$ 2,6\\
\hline
\multirow{ 2}{*}{LM-H} & F1 & 38,8 & \textbf{58,5} $\pm$ 4,7 & 58,4 $\pm$ 4,2 & 53,7  $\pm$ 3,0  & 52,5 $\pm$ 2,6 & \textbf{59,0} $\pm$ 3,1 & 56,6 $\pm$ 4,2 & 55,3 $\pm$ 3,2 &  52,6 $\pm$ 3,5 \\
& Acc. & 40,1 & \textbf{62,1} $\pm$ 2,1 & \textbf{62,3} $\pm$ 2,2 & 58,9 $\pm$ 2,2 & 57,4 $\pm$ 3,0 & 61,6 $\pm$ 1,6  & 61,1 $\pm$ 2,1 & 60,9 $\pm$ 1,7  &  58,6 $\pm$ 2,1\\
\hline
\multirow{ 2}{*}{Multiclass} & F1 & 24,2  & \textbf{52,4} $\pm$ 5,0 & 48,0 $\pm$ 2,8 & 43,6 $\pm$ 2,7 & 42,6 $\pm$ 2,4 & \textbf{48,5} $\pm$ 3,1 & 44,7 $\pm$ 3,0 & 44,1 $\pm$ 3,2  & 43,6 $\pm$ 3,9  \\
& Acc. & 32,3  & \textbf{53,9} $\pm$ 2,8 & 49,9 $\pm$ 2,4 & 46,5 $\pm$ 2,7 & 46,4 $\pm$ 3,0 & \textbf{50,4} $\pm$ 1,6 & 47,5 $\pm$ 2,6 & 46,8 $\pm$ 3,4  &  44,1 $\pm$ 3,4\\
\hline
\end{tabular}
\end{center}
\label{results_stress_level}
\end{table*}

\begin{table*}[htbp]
\caption{Root Mean Square Error (RMSE) for the regression of PHQ-9 depression score using the FCN model. Best and second best results and marked in bold face.}
\begin{center}
\begin{tabular}{|c|c|c|c|c|c|c|c|c|c|}
\hline
   & mean & all features & WiFi & GPS & Social & Phone log & Activity & Audio  &    Academic \\
\hline
PHQ-9 regression  & 4.4 & \textbf{4,1} $\pm$ 4,4 & \textbf{3,5} $\pm$ 3,4 & 4,9 $\pm$ 3,9 & 4,7 $\pm$ 4,3 & 5 $\pm$ 4,1 & 4,9 $\pm$ 4,7 & 5,1 $\pm$ 4,2  & 4,9 $\pm$ 4,1\\
\hline
\end{tabular}
\end{center}
\label{results_depression}
\end{table*}

\subsection{Modeling daily self-reported stress level}

We predict daily stress levels based on current (and past) passive data. 
We rescale the original stress data that comes in a 1-5 scale into 3 stress levels: below-median (low stress), median (medium stress) and above-median (high stress). We compute the median separately for each individual, as done in \cite{mikelsons2017towards}.  We test 3 different problem configurations: binary low versus high classification (L-H); binary low and medium versus high classification (LM-H); and multiclass classification (low/medium/high). We train two types of neural networks: a Fully Connected Network (FCN) that takes passive data from a single day as input; and a Recurrent Neural Network (LSTM) that takes data from 5 past days and predicts the stress of the last day. The FCN is composed by 3 layers (57, 35, 3), each one followed by ReLU activation and a dropout layer (0.35, 0.15, 0.15). The recurrent network is composed by a LSTM layer with 50 nodes and a dropout of 0.2, followed by a fully connected layer of 15 neurons with ReLU activation. We order chronologically the data, used the 80\% first days of the study for training and the last 20\% for testing. For this task, we also add an extra one-hot encoded variable indicating the user.
\subsection{Modeling depression score (PHQ-9)}

For predicting depression, we used the PHQ-9 (Patient Health Questionnaire), a self-report measure of depression symptom severity, with scores ranging from 0 to 27 (each item scored from 0 to 3). In total, 38 students applied for the PHQ-9 at the end of the study.  For this purpose, we created a FCN with Root Mean Square Error (RMSE) loss to predict the depression score. The FCN is composed by 3 layers (128 neurons) each one followed by ReLU activation function and a dropout layer of 0.3. We perform leave-one-out evaluation, training using 37 subjects and testing with one subject. 
The input data consists of aggregated features representing the mean and standard deviation of each feature over a period of two weeks leading up to the end of the study.

\section{Results and Discussion}

The accuracy and F1 score for stress level classification are reported in Table \ref{results_stress_level}. In turn, Table \ref{results_depression} reports the Root Mean Square Error (RMSE) for the depression score regression. 
We train each model with all $125$ features, as well as with each group of features separately. In all the cases we perform $50$ rounds of training and testing with different random initializations of the model. The tables report the mean and standard deviation results obtained per each performance measure. As a baseline, we include the accuracy and F1 score for a classifier that always predicts the most frequent stress level value for each student and RMSE for a regressor that always predicts the mean of all the PHQ-9 scores.

\textbf{Performance results for stress level classification}. The results in Table \ref{results_stress_level} show that, in general, the best performance is obtained when using all the features. 
Surprisingly, for the binary predictions the FCN model outperforms the LSTM, suggesting that using passive sensor data from the same day is more informative than incorporating data from the past 5 days. For the multi-class classification, the LSTM performs better than FCN, indicating that more context from past days is needed for a finer-grained classification problem.

\textbf{Performance results for PHQ-9 regression}. Table \ref{results_depression} shows that the model trained with mobility features embedded in WiFi data outperforms the model using all features. WiFi features, in particular, also exhibit a lower standard deviation, which supports the finding that these features have significant relevance for predicting depression. This can be attributed to the correlation between the time a person spends at home.
 
\textbf{Feature importance}. We observe that WiFi and Phone log features exhibit higher performance when used as individual group features, when compared to the other feature groups. WiFi data captures the mobility patterns of students within the experiment, while phone log features indirectly represent sleep patterns through indicators such as phone charging and time spent in the dark environment. These features also have lower standard deviations, indicating their consistency and relevance. On the contrary, social features show the worst performance overall, suggesting that metrics like the number of calls or SMS may not be highly influential for stress level prediction in this dataset. However, we notice that social features represent the smallest feature group, and their limited impact could be due to their lack of  information for stress prediction. The results suggest that mobility patterns and sleep habits have relevant information for stress prediction.

\textbf{Comparison of stress level classification with previous works}. Previous studies on stress level classification with the StudentLife dataset report results that are comparable with the results obtained by our approach. For example, for the binary stress level classification, In \cite{acikmese2019prediction}, an accuracy of 62.8\% is reported, while our model reaches from 60\% to almost 65\%. Kadri et al \cite{kadri2022hybrid} reports 93\% but they are using active data that can have a direct correlation with stress, so it is not comparable with our approach. In particular, the state-of-the-art results in multi class stress prediction with the StudentLife dataset  is obtained with a complex multitask architecture \cite{shaw2019personalized}, difficult to generalize to new unseen participants and using a reduced subset of students. Furthermore, we have observed a strong sensitivity to model initialization. For example, notice that the standard deviations range from $1,6$ to $3,9$ in the case of accuracy for the binary predictions. Unfortunately, none of these previous works has published their code, trained models, training-test data partitions or details on hyperparameters, which does not allow us to reproduce their results and perform a fair comparison. We encourage the practice of releasing code, so the community can fairly reproduce the reported results.

\textbf{Comparison of depression prediction with previous works.} Previous studies primarily focused on detecting depression using the Studentlife dataset, and they mainly relied on classification techniques \cite{colbaugh2020detecting}. However, our research takes a complementary approach by focusing on regression analysis. Our findings align with the results reported in \cite{saeb2016relationship}, which emphasize the significance of mobility features for detecting depression. Again, we also notice that none of these previous works released their code, trained models, or data partition details, making reproducibility and comparison not possible.

\textbf{Weaknesses}. In our study, the number of features differs depending on the group, and this might have an impact on the amount of information each group of features might encode. The encoded features are inspired by previous works, but other additional features could be computed from the raw data, which might also have an impact on the results.

\section{Conclusions}

In this study we use the StudentLife dataset to compare the contribution of various passive data features for stress level prediction and depression score regression. We found that different feature sets exhibited varying performance levels when tested with Neural Network models. WiFi features were particularly informative for stress and depression prediction, highlighting the importance of mobility patterns. Phone log features has relevance in stress level prediction, indicating the relation of sleep patterns and mental health states.


\section*{Ethical Impact Statement}
Our work focuses on evaluating the importance of different sensor data features in predicting depression and stress in students.  Our analysis aims to contribute the understanding of mental health states and enhancing interventions. However, we acknowledge the limitations of our study and emphasize the need for further research and validation before implementing it in real-world diagnosis. It is crucial to recognize the complexity of mental health disorders, which may not be fully reflected in the data collected from wearable devices or smartphones.
In our experiments, we use a publicly available and anonymized dataset where participants gave their consent for the collection of personal data. However, this dataset is very small and does not represent global student's conditions and general human behavior patterns. Therefore, larger, more balanced, and diverse datasets are necessary for future research. Furthermore, considerations such as privacy, data protection, and participant well-being should be carefully addressed when creating these datasets and training machine learning models. Lastly, machine learning models should not be seen as a replacement of the diagnosis provided by clinicians, but rather serve as a supportive tool.

\section*{Acknowledgments}

This work is partially supported by the Spanish Ministry of Science, Innovation and Universities RTI2018-095232-B-C22 to A.L and by Fundacion Bancaria La Caixa (ref. 2019-198898) to M.B. We also thank NVIDIA for their generous hardware donations.

\bibliographystyle{IEEEtran}
\bibliography{IEEEabrv,bibliography}

\end{document}